\pdfoutput=1

\documentclass[11pt]{article}

\usepackage{acl}

\usepackage{times}
\usepackage{latexsym}

\usepackage[T1]{fontenc}

\usepackage[utf8]{inputenc}

\usepackage{microtype}

\usepackage{amsmath}
\usepackage{pdfpages}
\usepackage{pgffor}
\usepackage{multirow}
\usepackage{booktabs}
\usepackage{color}
\usepackage{colortbl}
\usepackage{cleveref}
\crefformat{section}{\S#2#1#3} 

\definecolor{xing}{rgb}{1.0, 0.03, 0.0}

\title{Unsupervised Sign Language Translation and Generation}

\author{
Zhengsheng Guo$^{\flat}$ \quad Zhiwei He$^{\ddag}$ \quad Wenxiang Jiao\quad Xing Wang \quad Rui Wang$^{\ddag}$ \\
\bf Kehai Chen$^{\flat}$ \quad  Zhaopeng Tu \quad Yong Xu$^{\flat}$ \quad Min Zhang$^{\flat}$ \\
  $^{\flat}$Harbin Insititute of Technology at Shenzhen \\
  $^{\ddag}$ Shanghai Jiao Tong University 
}

\begin{document}
\maketitle
\begin{abstract}
Motivated by the success of unsupervised neural machine translation (UNMT), we
introduce an unsupervised sign language translation and generation network (USLNet), which learns from abundant single-modality (text and video) data without parallel sign language data. USLNet comprises two main components: single-modality reconstruction modules (text and video) that rebuild the input from its noisy version in the same modality and cross-modality back-translation modules (text-video-text and video-text-video) that reconstruct the input from its noisy version in the different modality using back-translation procedure.
Unlike the single-modality back-translation procedure in text-based UNMT, USLNet faces the cross-modality discrepancy in feature representation, in which the length and the feature dimension mismatch between text and video sequences.
We propose a sliding window method to address the issues of aligning variable-length text with video sequences.
To our knowledge, USLNet is the first unsupervised sign language translation and generation model capable of generating both natural language text and sign language video in a unified manner.
Experimental results on the BBC-Oxford Sign Language dataset (BOBSL) and Open-Domain American Sign Language dataset (OpenASL) reveal that USLNet achieves competitive results compared to supervised baseline models, indicating its effectiveness in sign language translation and generation.
\end{abstract}

\section{Introduction}

Sign language translation and generation  (SLTG) have emerged as essential tasks in facilitating communication between the deaf and hearing communities~\citep{phoenix}. Sign language translation involves the conversion of sign language videos into natural language, while sign language generation involves the generation of sign language videos from natural language.

Sign language translation and generation have achieved great progress in recent years. However, training an SLTG model requires a large parallel video-text corpus, which is known to be ineffective when the training data is insufficient~\citep{muller2022findings}. Furthermore, manual and professional sign language annotations are expensive and time-consuming. Inspired by the successes of unsupervised machine translation (UNMT)~\citep{artetxe2018unsupervised, lampleunsupervised} and unsupervised image-to-image translation~\citep{liu2017unsupervised}, we propose an unsupervised model for SLTG that does not rely on any parallel video-text corpus.

In this work, we propose an unsupervised SLTG network (USLNet), which learns from abundant single-modal (text and video) data without requiring any parallel sign language data. Similar to UNMT,  USLNet consists of the following components: the text reconstruction module (\cref{sec:text_reconstruction_module}) and the sign video reconstruction module (\cref{sec:sign_video_reconstruction_module}) that rebuild the input from its noisy version in the same modality, and cross-modality back-translation modules (\cref{sec:cross_modality_back_translation_module}) that reconstruct the input from its noisy version in the different modality using a back-translation procedure.

Unlike the single-modal back-translation in text-based UNMT, USLNet faces the challenge of cross-modal discrepancy. Sign and spoken languages exhibit distinct characteristics in terms of modality, structure, and expression. Sign language relies on visual gestures, facial expressions, and body movements to convey meaning, while spoken language depends on sequences of phonemes, words, and grammar rules~\citep{chen2022simple}. The cross-modal discrepancy in feature representation presents unique challenges for USLNet.

To address the cross-modal discrepancy in feature representation,  a common practice is to use a linear projection to map the representations from the single-modal representation to a shared multi-modal embedding space~\citep{radford2021learning}. This approach effectively bridges the gap between different feature representations, facilitating seamless integration of information and enhancing the overall performance of models in handling cross-modal translation tasks. In this work, we propose a sliding window method to address the issues of aligning the text with video sequences. 


To the best of our knowledge, USLNet is the first unsupervised SLTG model capable of generating both text and sign language video in a unified manner. Experimental results on the BBC-Oxford Sign Language dataset (BOBSL) ~\citep{bbc} and Open-Domain American Sign Language dataset (OpenASL)~\citep{openasl} reveal that USLNet achieves competitive results compared to the supervised baseline model ~\citep{sincan2023context, openasl} indicating its effectiveness in sign language translation and generation. 

Our contributions are summarized below:
\begin{enumerate}
    \item USLNet is the first unsupervised model for sign language translation and generation,  addressing the challenges of scarce high-quality parallel sign language resources.
    \item USLNet serves as a comprehensive and versatile model capable of performing both sign language translation and generation tasks efficiently  in a unified manner.
    \item USLNet demonstrates competitive performance compared to the previous supervised method on the BOBSL dataset.
\end{enumerate}

\section{Methodology}

The proposed framework in this study consists of four primary components: a text encoder, a text decoder, a video encoder, and a video decoder. As illustrated in Figure~\ref{fig:framework}, the USLNet framework encompasses four modules: \textbf{a text reconstruction module} (gray line in Figure~\ref{fig:framework}), \textbf{a sign video reconstruction module} (blue line in Figure~\ref{fig:framework}), \textbf{a text-video-text back-translation (T2V2T-BT) module} which initially translates input text into pseudo video (red line in Figure~\ref{fig:framework}) and subsequently back-translates pseudo video into text (yellow line in Figure~\ref{fig:framework})), and \textbf{a video-text-video back-translation (V2T2V-BT) module} which firstly translates input video into pseudo text (yellow line in Figure~\ref{fig:framework}) and then back-translates pseudo text into video (red line in Figure~\ref{fig:framework}). The latter two modules are considered cross-modality back-translation modules due to their utilization of the back-translation procedure. In this section, we will first describe each module and then introduce the training procedure.

\paragraph{Task Definition} 
We formally define the setting of unsupervised sign language translation and generation. Specifically, we aim to develop a USLNet that can effectively perform both sign language translation and generation tasks, utilizing the available text corpus  $\mathcal{T} = \{{\bf t}^i\}_{i=1}^{M}$, and sign language video corpus $\mathcal{V} = \{{\bf v}^j\}_{j=1}^{N}$, where $M$ and $N$ are the sizes of the text and video corpus, respectively.

\subsection{Text Reconstruction Module}
\label{sec:text_reconstruction_module}
\begin{figure*}[t!]
    \centering
    \includegraphics[height=0.40\textwidth]{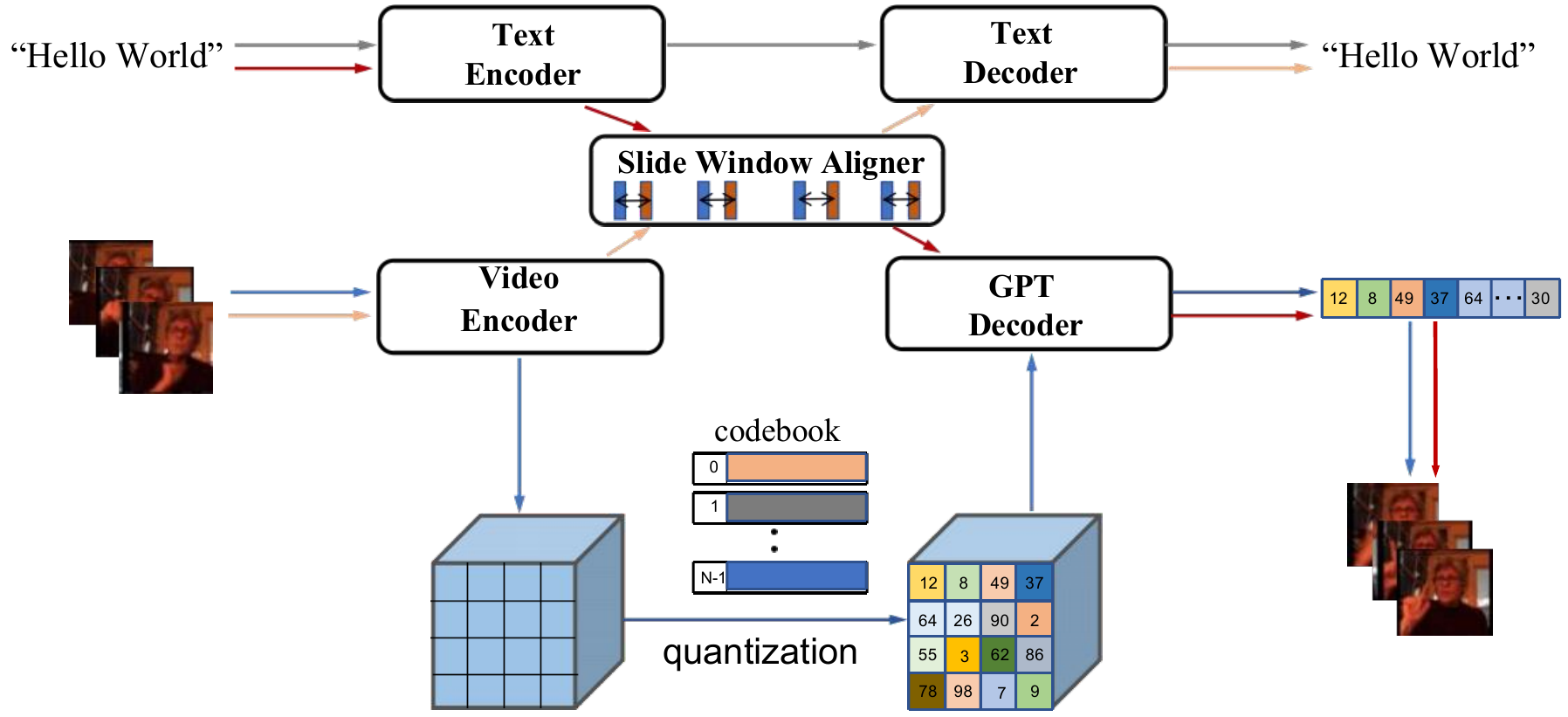}
          \caption{The overall framework of the proposed USLNet. The gray line denotes the text reconstruction procedure. The blue line denotes the video reconstruction procedure . The yellow line denotes the sign language translation procedure which translates video into the corresponding text. The red line denotes the sign language generation procedure which translates text into the corresponding video.   }
    \label{fig:framework}
\end{figure*}

As shown in Figure~\ref{fig:framework}, the text reconstruction module uses text encoder and  text decoder to reconstruct the original text from its corrupted version. Following the implementation by \citep{song2019mass}, we employ masked sequence-to-sequence learning to implement the text reconstruction. Specifically,  given an input text $\bf t$ = $\left(\bf t_{1}, \bf \ldots, \bf t_{n}\right)$ with $n$ words, we randomly mask out a sentence fragment  $\bf t ^{u:v}$ where $0 < u < v < n$ in the input text to construct the prediction sequence. The text encoder \textsc{Enc-text} is utilized to encode the masked sequence $\bf t ^{\setminus u:v}$, and the text decoder \textsc{Dec-text} is employed to predict the missing parts $\bf t ^{u:v}$. The log-likelihood serves as the optimization objective function:

\begin{equation}
    \mathcal{L}_{\text{text}} = \frac{1}{|\mathcal{T}|}\sum_{t \in \mathcal{T}}logP(\bf t ^{u:v} | \bf t ^{\setminus u:v})
\end{equation}

This task facilitates the model's learning of the underlying text structure and semantics while enhancing its capacity to manage noisy or incomplete inputs.

\begin{figure}[htbp]
	\centering
	\includegraphics[width=\columnwidth]{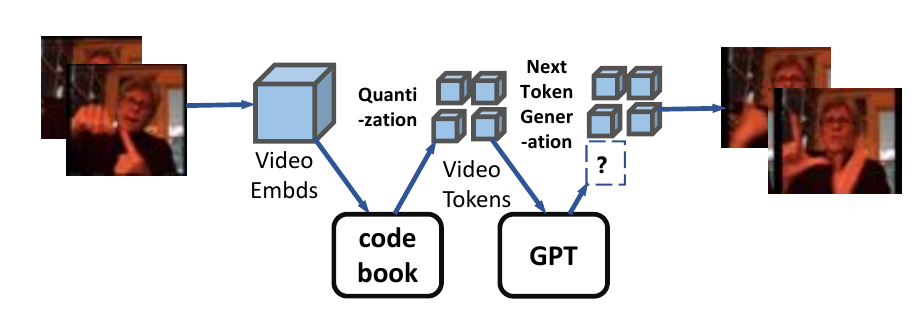}
	\caption{A figure describing sign video reconstruction module. This module is responsible for reconstructing the original video from the downsampled discrete latent representations of raw video data. In the quantization stage, the module transforms the video embeddings into discrete video tokens using a codebook. These video tokens are then input into GPT to generate the next visual token.}
	\label{fig:signRecons}
\end{figure}

\subsection{Sign Video Reconstruction Module}
\label{sec:sign_video_reconstruction_module}

Shown in Figure ~\ref{fig:signRecons}, the sign video reconstruction module  reconstructs the original video from the downsampled discrete latent representations of raw video data. In this work, we adopt the VideoGPT~\citep{videogpt} architecture to build the sign video reconstruction module. VideoGPT consists of two sequential stages, i.e., quantization and video sequence generation.

\paragraph{Quantization} VideoGPT employs 3D convolutions and transposed convolutions along with axial attention for the autoencoder in VQ-VAE, learning a downsampled set of discrete latents from raw pixels of the video frames. 

Specifically in the quantization stage, given an input video $\bf v = \left(v_{1}, \ldots, v_{n}\right)$ with $n$ pixels, the video encoder encodes the input $\bf v$ into video embeddings $\bf E_v =\left(E_{v_1}, \ldots, E_{v_{n}}\right) $, then $\bf E_v$ are discretized  by performing a nearest neighbors lookup in a codebook of embeddings  $\bf C=\left\{e_{i}\right\}_{i=1}^{N}$, as shown in Eq.(2).
 Next, $\bf E_v$ can be representated as discrete encodings $\bf E_{v}^{q}$ which consists of the nearest embedding indexs in codebook, shown in Eq.(3). Finally, video decoder learns to reconstruct the input  $\bf v$ from the quantized encodings.
\begin{equation} 
    \bf E_{v_{i}}=e_{k}, 
    \text { where } k=\operatorname{argmin}_{j}\left\|E_{v_{i}}-e_{j}\right\|_{2}
\end{equation}
\begin{multline}
   \bf E_{v} \rightarrow E_{v}^{q} = (k_1, \ldots, k_n), \\ \text{where} \quad k_{i}=  \operatorname{argmin}_{j}\left\|E_{v_{i}}-e_{j}\right\|_{2}
\end{multline}

The similarity between $\bf E_{v_{i}}$ and $\bf e_j$ serves as the optimization objective function:
\begin{equation}
   \mathcal{L_{\text{codebook}}} = \frac{1}{|\mathcal{C}|}\sum_{e_{j} \in \mathcal{C}} \left\|E_{v_{i}}-e_j\right\|_{2} 
\end{equation}

\paragraph{Video Sequence Generation} After quantization stage, the discrete video encodings $\bf E_{v}^{q}= (k_1, \ldots, k_n)$ are feed  into the GPT decoder, and generate the next video "word" $\bf k_{n+1}$. The similarity between autoregressively generated video $\bf v_{recon}$ and the original input video $\bf v$ serves as the optimization object function:

\begin{equation}
     \mathcal{L_{\text{video}}} = \frac{1}{|\mathcal{V}|}\sum_{v \in \mathcal{V}} \left\|v_{recon}-v\right\|_{2} 
\end{equation}

\subsection{Cross-modality Back-Translation Module}
\label{sec:cross_modality_back_translation_module}
The cross-modality back-translation module consists of two tasks: text-video-text back-translation (T2V2T-BT) and video-text-video back-translation (V2T2V-BT). In contrast to conventional back-translation~\citep{sennrich2016improving}, which utilizes the same modality, cross-modal back-translation encounters the challenge of addressing discrepancies between different modalities~\citep{ye2023cross}. Inspired by the recent work Visual-Language Mapper~\citep{chen2022simple}, we propose the implementation of a sliding window aligner to facilitate the mapping of cross-modal representations.

\paragraph{Sliding Window Aligner}
\begin{figure*}
	\centering
	\includegraphics[width=2\columnwidth]{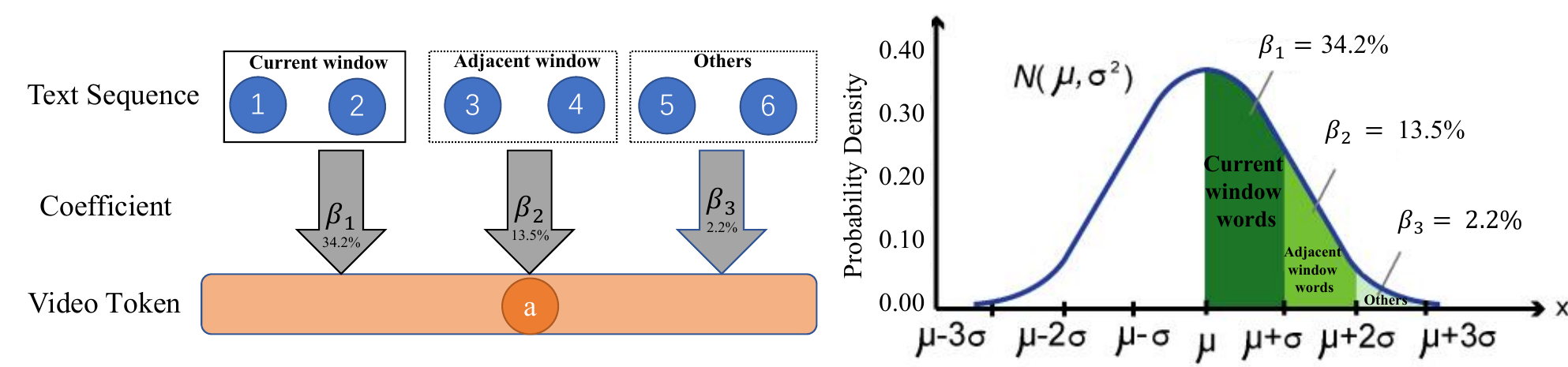}
   \caption{Left: A figure describing slide window aligner at step one. Right: Visualization of the
probability distribution (Gaussian distribution) that satisfies the weight coefficients of words in
different positions. At step one, we compute the first token "a" of pseudo video "sequence" by slide
window aligner.}
	\label{fig:slide}
\end{figure*}


The sliding window aligner is proposed to address the discrepancies between text and video modal representations. Specifically, two primary distinctions between text and video representation sequences are hidden dimensions and sequence length differences. Considering these differences, the aligner consists of two components: \emph{length mapper} $\bf M^L$ and \emph{dimension mapper} $\bf M^D$. Considering different back-translation directions (V2T2V and T2V2T), dimension mappers include text-to-video mapper $\bf M_{T\rightarrow V}^D$ and video-to-text mapper $\bf M_{V\rightarrow T}^D$.

Given the text  encoder output $\bf E_t$, the text decoder input $\bf D_t$, the codebook reconstructed video embedding $\bf E_v$ and video GPT input $\bf D_v$, the feature dimension transformation procedure are as follows:

\begin{gather}
   \bf D_v = M^L(M_{T\rightarrow V}^D(E_t)) \\
    \bf D_t = M^L(M_{V\rightarrow T}^D(E_v)) 
\end{gather}


Aiming to solve the length discrepancy, we design \textbf{length mapper} $\bf M^L$ method, which uses the sliding window method. According to ~\citep{BSLlinguistics}, signing is particularly influenced by English word order when the signers sign while translating from a text. In the context of British Sign Language, presenters may adhere to a more English-like word order. Drawing upon this linguistic understanding, we propose a method wherein the source sequence is partitioned into distinct windows, allowing each word in the target sequence to align more closely with its corresponding source window. 

 Taking text-to-video for example, supposed that input text sequence $\bf t$ = $\left(\bf t_{1}, \bf \ldots, \bf t_{m}\right)$ with m words, video sequence $\bf v = \left(v_{1}, \ldots, v_{n}\right)$ with n frames and $\bf m>n$,  the sliding window method, Length Mapper $\bf M^L$ which can be described as follows:

\begin{equation}
   \bf v_i = \sum\limits_{i=1}^{n}\alpha_i t_i 
\end{equation}
\begin{multline}
    \begin{bmatrix}
        \bf \alpha_1 &  \dots & \bf \alpha_n
    \end{bmatrix} =   \text{softmax}\left(
    \begin{bmatrix}
        \bf \beta_1 &  \dots & \bf \beta_n
    \end{bmatrix}\right)
\end{multline}
\begin{multline}  
    \bf \beta_i  \in \begin{Bmatrix}
 (p(\mu +\sigma ), & p(\mu)] ,            &  i  \in  W_{c} \\
  (p(\mu +2\sigma ), & p(\mu+\sigma )] , &  i  \in W_{a} \\
  (p(\mu +3\sigma ),& p(\mu+2\sigma )]  , & i \in W_{o}
\end{Bmatrix} 
\end{multline}

Show Eq.(8),  every video word accept all text words' information. However, each word in the target sequence aligns more closely with its corresponding window. For example, the beginning video frames conveys more information about the first some text words. Specifically, weight coefficient $[\bf \alpha_1,  \alpha_2, \ldots, \alpha_n]$ comes from $ X = [\bf \beta_1,  \beta_2, \ldots, \beta_n]$. X follows  a Gaussian  distribution $N(\mu, \sigma^2)$. The value of $\bf \beta_i$ depends on where token i is and is divided into three probability intervals $(p(\cdot), p(\cdot)]$, shown in Eq.(10). $W_c, W_a, W_o$ represent distinct positional intervals, namely the current window, adjacent window, and other positions. The value of token $\bf \beta_i$ exhibits an upward trend as its proximity to the current window increases. In the case where token $i$ falls within the bounds of the current window $W_c$, the weight coefficient is assigned to the highest intervals. 

\begin{figure*}
	\centering
\includegraphics[width=2.15\columnwidth]{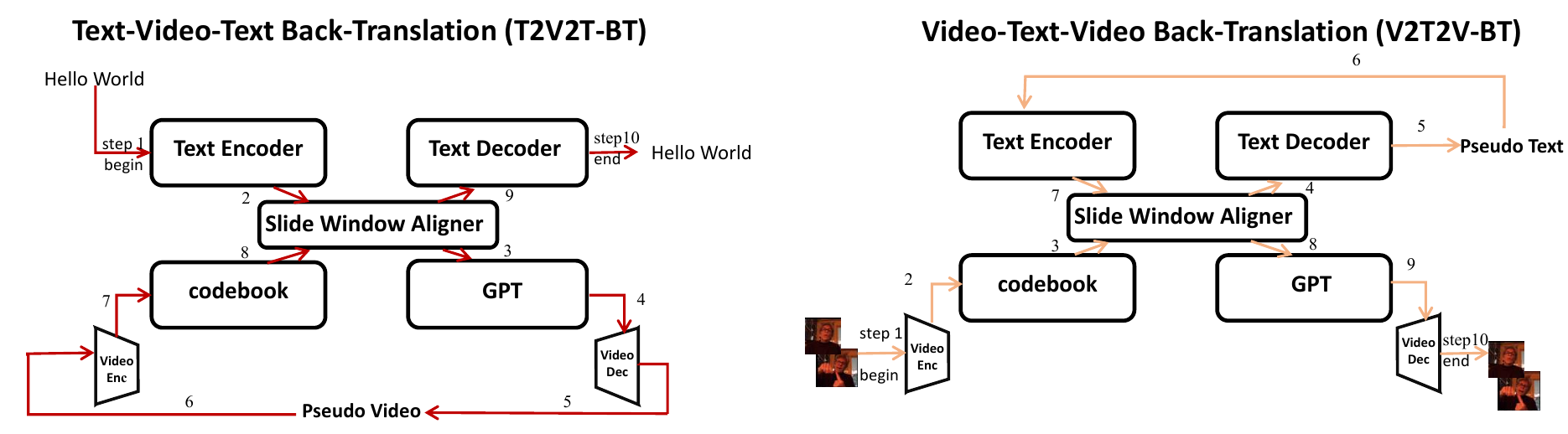}
	\caption{A figure describing the procedure of cross-modality back-translation. The left sub-figure depicts the Text-Video-Text Back-Translation (T2V2T-BT) procedure, while the right sub-figure showcases the Video-Text-Video Back-Translation (V2T2V-BT) procedure. Each sub-figure provides a step-by-step description of the respective back-translation process. The numbers assigned next to the arrows indicate the sequential order of the steps. For instance, "2" signifies that the step is the second step in the procedure.}
	\label{fig:cross}
\end{figure*}

For example, supposed text has 6 words $\bf t=(t_1, \ldots, t_6)$ and video has 4 frames $\bf v = (v_a, v_b, v_c, v_d)$. The window size can be computed as $\lceil 6/4 \rceil = 2$. As Figure ~\ref{fig:slide} has shown, when generating the first video token $v_a$, it incorporates information from all text tokens while placing the highest weight coefficient $\beta_1$ on the first few text words $W_c$. Meanwhile, the value of token $\mathbf{\beta_i}$ exhibits a declining trend as its proximity to the current window diminishes ($\beta_1 > \beta_2 > \beta_3$). 

We introduce \textbf{dimension mapper} $\bf M^D$  to address the differences in hidden dimensions of different modalities. For example, $\bf M_{T\rightarrow V}^D(E_t)$ transposes text embeddings' hidden dimensions into video embeddings' hidden dimensions, facilitating the integration and alignment of textual and visual information for improved multimodal tasks.

\paragraph{Cross-Modality Back-Translation}

The T2V2T-BT translates a given text sequence into a sign video, followed by translating the generated sign video back into text, shown in Figure ~\ref{fig:cross}. The objective of T2V2T-BT is to ensure consistency between the generated text and the original text while accurately translating the video back into the original text. This task assists the model in capturing the semantic and visual correspondence between text and video modalities and comprehending the input data's underlying structure and temporal dynamics. The similarity between back-translated text $\bf t_{BT}$ and the original input text $\bf t$ serves as the optimization object function:

\begin{equation}
     \mathcal{L_{\text{T2V2T}}} = \frac{1}{|\mathcal{T}|}\sum_{t \in \mathcal{T}} \left\|t_{BT}-t\right\|_{2} 
\end{equation}

Similarly, the V2T2V-BT task requires the model to translate a given video into its corresponding text description, and then translate the generated text back into a video, using the original video as a reference, shown in Figure ~\ref{fig:cross}. The similarity between back-translated video $\bf v_{BT}$ and the original input video $\bf v$ serves as the optimization object function:

\begin{equation}
     \mathcal{L_{\text{V2T2V}}} = \frac{1}{|\mathcal{V}|}\sum_{v \in \mathcal{V}} \left\|v_{BT}-v\right\|_{2} 
\end{equation}

Overall, the cross-modality back-translation module of our proposed USLNet aims to improve the model's ability to translate between text and video modalities in an unsupervised manner, by learning a consistent and meaningful mapping between the two modalities.

 \subsection{Unsupervised Joint Training}

The training objective of USLNet combines the aforementioned loss terms, enabling joint optimization of the text and video networks. The losses $L_{text}$ and $L_{video}$ encourage the generator network to reconstruct the input from its noisy version within the same modality, while the losses $L_{T2V2T}$ and $L_{V2T2V}$ encourage USLNet to reconstruct the input from its noisy version across different modalities. This joint training approach empowers USLNet to not only exhibit strong single-modality generation capabilities in text and video but also acquire cross-modality mapping abilities.
\begin{multline}
    \mathcal{L_{\text{overall}}} = \alpha_1\mathcal{L_{\text{text}}} + \alpha_2\mathcal{L_{\text{codebook}}} + \alpha_3\mathcal{L_{\text{video}}} + \\
    \alpha_4\mathcal{L_{\text{T2V2T}}} + 
    \alpha_5\mathcal{L_{\text{V2T2V}}} 
\end{multline}

\section{Experiment}

\paragraph{Dataset}
We conduct a comprehensive evaluation of our approach using two distinct large-scale sign language translation datasets. \textbf{BBC-Oxford British Sign Language Dataset (BOBSL)}~\citep{bbc} is the largest existing video collection of British sign language (BSL).  It comprises 1,004K, 20K, and 168K samples in the train, dev, and test sets, respectively. The vocabulary size amounts to 78K, with an out-of-vocabulary (OOV) size of 4.8K in the test set. The second dataset we utilize is \textbf{OpenASL} ~\citep{openasl}, an expansive American Sign Language (ASL) - English dataset collected from various online video platforms. OpenASL boasts an impressive collection of 288 hours of ASL videos across multiple domains, featuring over 200 signers. 

\paragraph{Metric}
The evaluation of USLNet comprises sign language translation (SLT) and sign language generation (SLG). For SLT task, we adopt the BLEU~\citep{bleu} as the evaluation metric for the sign language translation. For SLG, we follow UNMT~\citep{lampleunsupervised} to utilize back-translation BLEU to assess the performance. Specifically, we back-translate the generated sign language video and use the input text as the reference to compute the BLEU score. Additionally,, we adopt Frechet Video Distance (FVD)~\citep{fvd} scores to evaluate the quality of generated video.

\paragraph{Model} 
The USLNet incorporates the MASS~\citep{song2019mass} architecture as the text model backbone and VideoGPT~\citep{videogpt} as the video model backbone. For the text model, we set the encoder and decoder layers to 6, and the hidden dimension to 1024. As for the video model, we build the VideoGPT with 8 layers and 6 heads,with a hidden dimension of 576. For the codebook, we set it with 2048 codes, wherein each code represents a feature tensor with a 256-dimensional. The training process comprises two stages: pre-training and unsupervised training. Firstly, we perform continued pre-training using the pre-trained MASS model~\citep{song2019mass} on the text portion of the dataset. Then, we train the VideoGPT model ~\citep{videogpt} on the sign language video component of the dataset. Finally, we utilize the pre-trained MASS and VideoGPT models to initialize the USLNet and conduct unsupervised joint training, as described in Section 2.4. We train the whole network  with a learning rate of 1e-3. Moreover, we use greedy decoding in evaluation procedure.

 \section{Results and Discussion}

 \subsection{Main Result}
 \paragraph{Sign Language Translation}

In Table 1, we present a comparative analysis between our approach and state-of-the-art methods for SLT on the BOBSL and OpenASL dataset. 

For unsupervised-based methods, given the fact that USLNet is the first unsupervised SLT method and BOBSL and openasl has no complete sentence-level gloss annotations datasets~\citep{bbc, openasl, lin2023gloss}, USLNet w/o, joint training is used to be unsupervised baseline. We observe an approximate improvement of 0.1 BLEU-4 on the BOBSL test set and 1.2 BLEU-4 on the OpenASL dataset. More qualitive results and analysis can be seen in Appendix A.1.

To ensure a fair evaluation of USLNet's effectiveness, we also present results for USLNet (S) , which represents USLNet in supervised settings, and USLNet (U+S) , where USLNet undergoes unsupervised training followed by supervised fine-tuning. We compare USLNet's performance in supervised settings against previous state-of-the-art methods. Remarkably, it is observed that USLNet attains new state-of-the-art (SOTA) performance on the BOBSL dataset, while also exhibiting competitive results on the OpenASL dataset. Importantly, USLNet (U+S)  outperforms both USLNet and USLNet (S) in both the BOBSL and OpenASL datasets, underscoring the effectiveness of unsupervised training in enhancing the representation of the SLT system.

\begin{table*}[h]
\centering
\begin{tabular}{c ccc ccc}
\toprule
\multirow{3}{*}{\bf Method} & \multicolumn{3}{c}{\bf BOBSL } & \multicolumn{3}{c}{\bf OpenASL } \\
\cmidrule(lr){2-4} \cmidrule(lr){5-7} 
& \bf Dev  & \multicolumn{2}{c}{\bf Test} & \bf Dev  & \multicolumn{2}{c}{\bf Test}  \\
& B@1$\uparrow$  & B@1$\uparrow$  & B@4$\uparrow$ & B@1$\uparrow$  & B@1$\uparrow$  & B@4$\uparrow$  \\
\hline
\rowcolor{gray!25}  \multicolumn{7}{c}{\textbf{Supervised Approach} } \\

Transformer~\cite{bbc}                   & -- & 12.78  & 1.00   & -- & -- & --   \\
Context-Transformer~\cite{sincan2023context} & 18.80 & 17.71 & 1.27   & -- & -- & --  \\ 
Conv-GRU~\citep{openasl} & -- & -- & -- & 16.72 & 16.11 & 4.58 \\
Transformer~\citep{openasl}  & -- & -- & -- & 20.10 & 20.92 & 6.72\\
USLNet (\textbf{S}) & 19.60  &15.50 & 1.00    & 15.40  & 16.90 & 4.30\\
USLNet (\textbf{U+S}) & 24.60  & 27.00 & 1.40 & 19.30  & 20.90 & 6.30\\
\hline
\rowcolor{gray!25}  \multicolumn{7}{c}{\textbf{Unsupervised Approach} } \\
USLNet  w/o.  joint training  & 1.40 & 1.50 & 0.00 & 1.60 & 1.30 & 0.00 \\
USLNet  w. joint training &  17.30  & 21.30 & 0.10  &  14.50  & 12.40 & 1.20 \\

\bottomrule
\end{tabular}
\caption{Sign language translation performance in terms of BLEU on BOBSL and OpenASL test set. B@1 and denotes BLEU-1 and BLEU-4, respectively. \textbf{S} represents supervised settings; \textbf{U+S} represents firstly unsupervised training and then supervised fine-tuning.}
\label{tab:slt}
\end{table*}

 \paragraph{Sign Language Generation}
Since there are no existing results for sign language generation on the BOBSL dataset, we compare the use of unsupervised joint training in USLNet. As shown in Table 2, the unsupervised joint training in USLNet yields improvements in terms of back-translation BLEU  and FVD scores, demonstrating the effectiveness of USLNet. More qualitive results can be seen in Appendix A.6.

\begin{table*}[htbp]
    \centering
    \resizebox{\linewidth}{!}{
    \begin{tabular}{c ccccc  ccccc}
        \toprule
       \multirow{3}{*}{\bf Method} &  \multicolumn{5}{c}{\bf BOBSL} &  \multicolumn{5}{c}{\bf OpenASL} \\ 
       \cmidrule(lr){2-6} \cmidrule(lr){7-11} 
       & \multicolumn{2}{c}{\bf Dev } &  \multicolumn{3}{c}{\bf Test}  & \multicolumn{2}{c}{\bf Dev } &  \multicolumn{3}{c}{\bf Test}\\
       &  B@1$\uparrow$ & FVD $\downarrow$  & B@1$\uparrow$  & B@4 $\uparrow$ & FVD $\downarrow$ &  B@1$\uparrow$ & FVD $\downarrow$  & B@1$\uparrow$  & B@4 $\uparrow$ & FVD $\downarrow$ \\
        \hline
       USLNet-P    & 0.50   & 892.8 &  0.70 & 0.00  & 872.7  & 1.50   & 886.4 &  1.30 & 0.00  & 890.2 \\
       USLNet   & 20.90 & 402.8 & 22.70 & 0.20 & 389.2 & 19.40   & 400.2 &  21.30 & 7.20  & 390.5 \\
        \bottomrule
    \end{tabular}}
    \caption{Sign language generation performance in terms of back-translation BLEU and  Frechet Video Distance (FVD) on BOBSL and OpenASL dataset.  B@1 and denotes BLEU-1 and BLEU-4, respectively. USLNet-P is the comparison baseline, representing USLNet  w/o.  joint training. USLNet represenents USLNet  w.  joint training. }
    \label{tab:slg}
\end{table*}

\begin{figure}[h]
	\centering
	\includegraphics[width=\columnwidth]{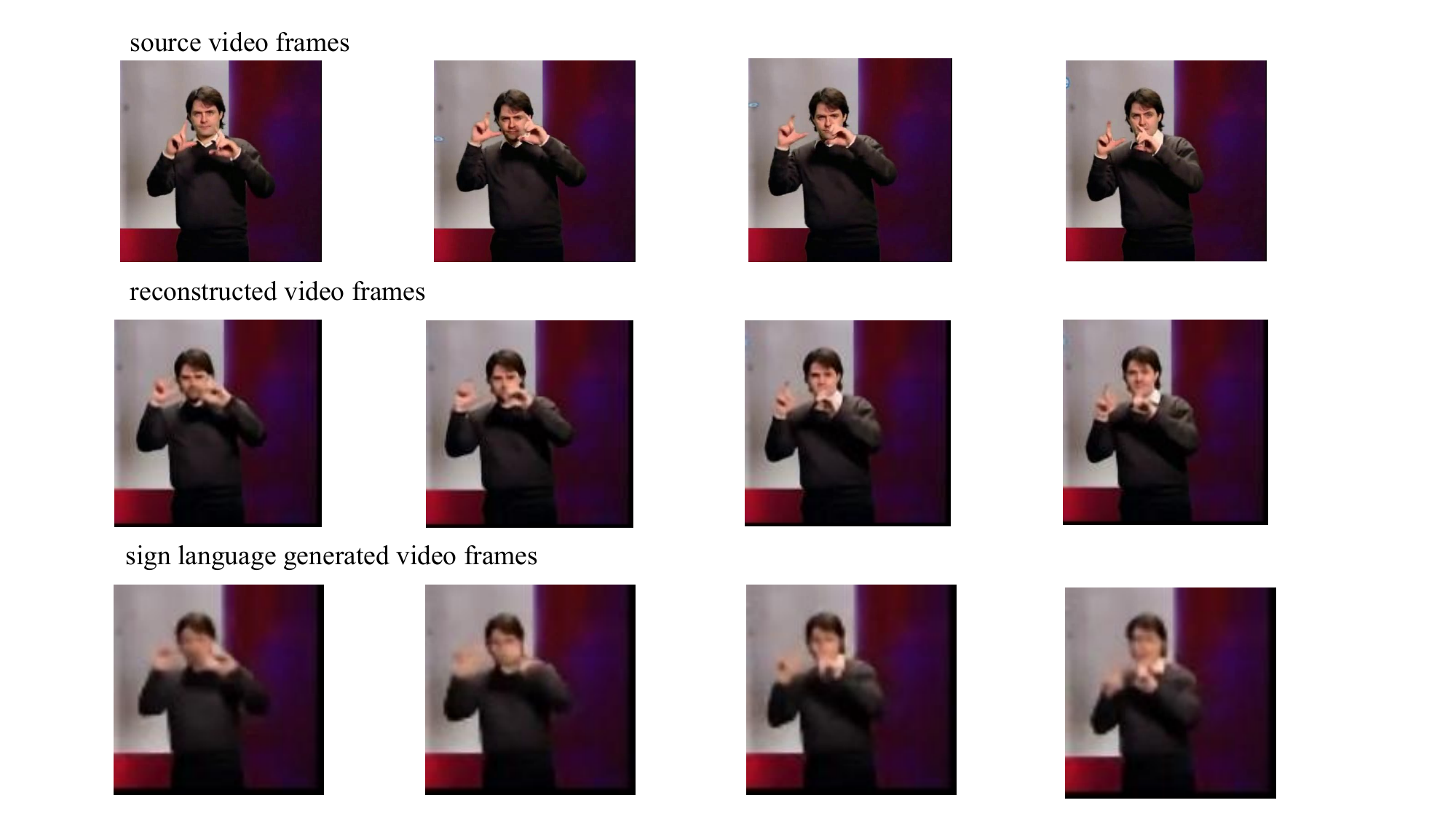}
	\caption{Case study of UnSLNet on BOBSL for sign language generation task. Examples are from test set.}
	\label{fig:{pic/model}}
\end{figure}

\subsection{Analysis}

In this section, we aim to gain a deeper understanding of the improvements achieved by USLNet. To achieve this, we evaluate the effectiveness of the proposed novel sliding window aligner from two perspectives: order consistency and slider comparision. 



\paragraph{Order Validation}Video and glosses are monotonically aligned. We hypothesis that video and text are roughly aligned. To verify this,  we must first obtain the golden sign order. Because OpenASL don't have gloss annotation in train set ~\citep{openasl}, we only verify it in BOBSL. Moreover, BOBSL does not have human-evaluated sentence-level glosses annotations, we utilized the automatic gloss annotation released in ~\citep{momeni2022automatic}. This gloss annotation consists of word-level annotations, presented as [video
name, global time, gloss, source, confidence]. We converted them into sentence-level annotations and assessed the consistency between the gloss (sign) and text orders.  From Table ~\ref{tab:orderValidation}, we can see the hypothesis that video and text are roughly aligned in BOBSL dataset.
\begin{table}[h]
   \centering
    \setlength{\tabcolsep}{1.5pt}
    \resizebox{\linewidth}{!}{
    \begin{tabular}{l | c }
       \toprule
          &  Proportion  \\
        \midrule
        Strictly Consistency & 0.83  \\
         Consistency with two gloss in disorder	  & 0.87  \\ 
       Consistency with three gloss in disorder	 & 0.91 \\
        \bottomrule
    \end{tabular}}
    \caption{Validation between sign(gloss) and text order consistency for BOBSL.}
    \label{tab:orderValidation}
\end{table}

\paragraph{Different Alignment Networks} To further explore the advantages of the proposed sliding window aligner (soft connection) , we have designed two comparison aligner networks, altering only the length mapper component $\bf M^L$. The first network is pooling, where the text sequence is padded to a fixed length and a linear network maps it to the video sequence length. The second network is the sliding window aligner with a hard connection, also utilizing a sliding window mechanism. However, $\alpha_{i}$  in Eq(9) is non-zero only if tokens are in the current window, indicating that it conveys information exclusively from tokens in the current window. As demonstrated in Table ~\ref{tab:main_German}, our method achieves the best performance. Moreover, different alignment networks for SLG can be seen in Appendix A.2.

    \begin{table}[!htbp]
    \centering
    \setlength{\tabcolsep}{2pt}
    \begin{tabular}{l | c  | cc}
        \toprule
         \multirow{2}{*}{\bf Method}  &  \multicolumn{1}{c|}{\bf Dev } &  \multicolumn{2}{c}{\bf Test} \\
        & B@1$\uparrow$   & B@1$\uparrow$  & B@4$\uparrow$  \\
        \midrule
        Pooling & 10.70 &  12.00  & 0.00 \\
        Sliding Window Aligner & 15.50  &  17.10 & 0.00   \\ 
        (hard connection) &&& \\
        Sliding Window Aligner & 17.30 & 21.30 & 0.10 \\
        (soft connection) &&& \\
        \bottomrule
    \end{tabular}
     \caption{Sign language translation results of USLNet with different cross-modality mappers on BOBSL. B@1 and denotes BLEU-1 and BLEU-4, respectively.}
    \label{tab:main_German}
\end{table} 

\paragraph{Comparision between BOBSL and WMT}
USLNet's performance on the BOBSL dataset is inadequate, similar to the performance observed on the WMT SLT task dataset where the state-of-the-art results showed low performance with  a BLEU-4 score of 0.56 ~\citep{WMT2022findings}. Our investigation revealed that the BOBSL dataset presents comparable difficulties to the WMT dataset. Notably, the BOBSL dataset possesses a substantially larger vocabulary of 72,000 words, compared to the WMT dataset's vocabulary of 22,000 words. 

\subsection{Ablation Study}
We conduct our ablation studies on the BOBSL dataset, evaluating the SLT BLEU-1 score on the
development set. 

\paragraph{Adjust Data Distribution} The transformation of un-parallel video and text data into parallel video and text data, employed in an unsupervised manner, has been demonstrated to significantly improve SLT  (+5.60 BLEU-1 score). 

\paragraph{Explore Different Freezing Strategy} 
Inspired by ~\citep{sltunet}, we we compare various freezing strategies by evaluating their impact on the performance of SLT.Our experimental results demonstrate that freezing video encoder can improve SLT effects (+2.10 BLEU-1 socore). 


\begin{table}[!htbp]
\centering
 \setlength{\tabcolsep}{1pt}
\begin{tabular}{ccc}
\toprule
ID & System & SLT B@1$\uparrow$ \\
\midrule
1 & Baseline & 3.20  \\
1.1 & 1+more text data & 9.60 \\
\hline

\rowcolor{gray!25}  \multicolumn{3}{c}{Explore Multi-Task Learning}   \\
2.1 & 1.1+ remove & \\
& text reconstruction at training & 5.40  \\
2.2 & 1.1+ remove & \\
& video reconstruction at training & 8.30 \\
2.3 & 1.1+remove cross-modality \\
& Back-Translation at training & 0.70 \\

\hline
\rowcolor{gray!25}  \multicolumn{3}{c}{Adjust Data Distribution} \\
3 & 1.1+ 1M parallel video and text  & \\
& for unsupervised training & 15.20  \\

\hline
\rowcolor{gray!25}  \multicolumn{3}{c}{Explore Different Freezing Strategy} \\
4.1 & 3+ freeze video decoder & 10.80 \\
4.2 & 3+ freeze text encoder & 12.20 \\
4.3 & 3+ freeze text decoder & 12.60 \\
4.1 & 3+ freeze video encoder  & 17.30  \\
\bottomrule
\end{tabular}
\caption{Ablation study of USLNet on sign language translation(SLT) on the BOBSL dev set.}
\end{table}

\section{Related Work}

\paragraph{Sign Language Translation} 
SLT involves translating sign language videos into text ~\citep{neural}. Previous SLT methods can be categorized into two groups: those focusing on enhancing visual encoder representation ~\citep{yin2021including, zhou2021spatial, yin2020better, kan2022sign}, and those aiming to improve text decoder quality ~\citep{signTransformer, chen2022simple, ye2023cross, angelova2022using, he-etal-2022-tencent, he2023exploring, ye2022scaling, signBT}. For large-scale SLT datasets like BOBSL and openASL, ~\citet{bbc} utilizes a standard transformer model, while ~\citet{sincan2023context} proposes a context-based approach to enhance quality. Additionally, ~\citet{openasl} incorporates pre-training and local feature modeling for capturing sign language features. To the best of our knowledge, our work represents the first exploration of unsupervised methods in the SLT domain.

\paragraph{Sign Language Generation} 

Sign language generation focuses on generating highly reliable sign language videos ~\citep{bragg2019sign, cox2002tessa}.  Previous research predominantly relied on high-quality parallel sign language video and text corpora~\citep{glauert2006vanessa,cox2002tessa, modelIntensity}. In our work, we aim to explore an unsupervised approach ~\citep{lampleunsupervised, artetxe2018unsupervised, he-etal-2022-bridging} that leverages unlabeled data for training the first SLG model. 

\section{Conclusion}



In this paper, we present an unsupervised sign language translation and generation network, USLNet. USLNet is the first bi-directional (translation/generation) sign language approach  trained in unsupervised manner. Experimental results on the large-scale sign dataset such as BOBSL and OpenASL reveal that USLNet achieves competitive performance compared to the supervised approach.



\section{Limitations}
Our USLNet for unsupervised sign language translation and generation has the following limitations:

\begin{itemize}
    \item \textbf{Performance on sign language translation and generation}:
As the pioneering unsupervised Sign Language Translation and Generation (SLTG) model, we acknowledge that USLNet's performance is not flawless and further advancements are needed, particularly in the realm of large-scale sign language. We recognize the significance of ongoing breakthroughs required to enhance USLNet's capabilities in this domain. 
    \item \textbf{Model Structure}:
USLNet has been designed with the objective of exploring a unified model that is capable of both sign language translation and generation. To achieve this, USLNet adopts a twin tower model, comprising separate components for text and video processing. Additionally, to treat videos as sequences, we have incorporated a video quantization model. These factors contribute to the complexity of the USLNet model, which necessitates substantial resources for training.
   
\end{itemize}


\bibliography{acl_latex}

\begin{thebibliography}{55}
\expandafter\ifx\csname natexlab\endcsname\relax\def\natexlab#1{#1}\fi

\bibitem[{Albanie et~al.(2021)Albanie, Varol, Momeni, Bull, Afouras, Chowdhury,
  Fox, Woll, Cooper, McParland et~al.}]{bbc}
Samuel Albanie, G{\"u}l Varol, Liliane Momeni, Hannah Bull, Triantafyllos
  Afouras, Himel Chowdhury, Neil Fox, Bencie Woll, Rob Cooper, Andrew
  McParland, et~al. 2021.
\newblock Bbc-oxford british sign language dataset.
\newblock \emph{arXiv preprint arXiv:2111.03635}.

\bibitem[{Angelova et~al.(2022{\natexlab{a}})Angelova, Avramidis, and
  M{\"o}ller}]{phoenix}
Galina Angelova, Eleftherios Avramidis, and Sebastian M{\"o}ller.
  2022{\natexlab{a}}.
\newblock Using neural machine translation methods for sign language
  translation.
\newblock In \emph{Proceedings of the 60th Annual Meeting of the Association
  for Computational Linguistics: Student Research Workshop}, pages 273--284.

\bibitem[{Angelova et~al.(2022{\natexlab{b}})Angelova, Avramidis, and
  M{\"o}ller}]{angelova2022using}
Galina Angelova, Eleftherios Avramidis, and Sebastian M{\"o}ller.
  2022{\natexlab{b}}.
\newblock Using neural machine translation methods for sign language
  translation.
\newblock In \emph{Proceedings of the 60th Annual Meeting of the Association
  for Computational Linguistics: Student Research Workshop}, pages 273--284.

\bibitem[{Artetxe et~al.(2018)Artetxe, Labaka, Agirre, and
  Cho}]{artetxe2018unsupervised}
Mikel Artetxe, Gorka Labaka, Eneko Agirre, and Kyunghyun Cho. 2018.
\newblock Unsupervised neural machine translation.
\newblock In \emph{6th International Conference on Learning Representations,
  ICLR 2018}.

\bibitem[{Borsos et~al.(2023)Borsos, Marinier, Vincent, Kharitonov, Pietquin,
  Sharifi, Roblek, Teboul, Grangier, Tagliasacchi et~al.}]{aligner10}
Zal{\'a}n Borsos, Rapha{\"e}l Marinier, Damien Vincent, Eugene Kharitonov,
  Olivier Pietquin, Matt Sharifi, Dominik Roblek, Olivier Teboul, David
  Grangier, Marco Tagliasacchi, et~al. 2023.
\newblock Audiolm: a language modeling approach to audio generation.
\newblock \emph{IEEE/ACM Transactions on Audio, Speech, and Language
  Processing}.

\bibitem[{Bragg et~al.(2019)Bragg, Koller, Bellard, Berke, Boudreault,
  Braffort, Caselli, Huenerfauth, Kacorri, Verhoef et~al.}]{bragg2019sign}
Danielle Bragg, Oscar Koller, Mary Bellard, Larwan Berke, Patrick Boudreault,
  Annelies Braffort, Naomi Caselli, Matt Huenerfauth, Hernisa Kacorri, Tessa
  Verhoef, et~al. 2019.
\newblock Sign language recognition, generation, and translation: An
  interdisciplinary perspective.
\newblock In \emph{Proceedings of the 21st International ACM SIGACCESS
  Conference on Computers and Accessibility}, pages 16--31.

\bibitem[{Camgoz et~al.(2018)Camgoz, Hadfield, Koller, Ney, and
  Bowden}]{neural}
Necati~Cihan Camgoz, Simon Hadfield, Oscar Koller, Hermann Ney, and Richard
  Bowden. 2018.
\newblock Neural sign language translation.
\newblock In \emph{Proceedings of the IEEE conference on computer vision and
  pattern recognition}, pages 7784--7793.

\bibitem[{Camgoz et~al.(2020)Camgoz, Koller, Hadfield, and
  Bowden}]{signTransformer}
Necati~Cihan Camgoz, Oscar Koller, Simon Hadfield, and Richard Bowden. 2020.
\newblock Sign language transformers: Joint end-to-end sign language
  recognition and translation.
\newblock In \emph{Proceedings of the IEEE/CVF conference on computer vision
  and pattern recognition}, pages 10023--10033.

\bibitem[{Chen et~al.(2022)Chen, Wei, Sun, Wu, and Lin}]{chen2022simple}
Yutong Chen, Fangyun Wei, Xiao Sun, Zhirong Wu, and Stephen Lin. 2022.
\newblock A simple multi-modality transfer learning baseline for sign language
  translation.
\newblock In \emph{CVPR}.

\bibitem[{Cox et~al.(2002)Cox, Lincoln, Tryggvason, Nakisa, Wells, Tutt, and
  Abbott}]{cox2002tessa}
Stephen Cox, Michael Lincoln, Judy Tryggvason, Melanie Nakisa, Mark Wells,
  Marcus Tutt, and Sanja Abbott. 2002.
\newblock Tessa, a system to aid communication with deaf people.
\newblock In \emph{Proceedings of the fifth international ACM conference on
  Assistive technologies}, pages 205--212.

\bibitem[{Glauert et~al.(2006{\natexlab{a}})Glauert, Elliott, Cox, Tryggvason,
  and Sheard}]{glauert2006vanessa}
John~RW Glauert, Ralph Elliott, Stephen~J Cox, Judy Tryggvason, and Mary
  Sheard. 2006{\natexlab{a}}.
\newblock Vanessa--a system for communication between deaf and hearing people.
\newblock \emph{Technology and disability}, 18(4):207--216.

\bibitem[{Glauert et~al.(2006{\natexlab{b}})Glauert, Elliott, Cox, Tryggvason,
  and Sheard}]{aligner1}
John~RW Glauert, Ralph Elliott, Stephen~J Cox, Judy Tryggvason, and Mary
  Sheard. 2006{\natexlab{b}}.
\newblock Vanessa--a system for communication between deaf and hearing people.
\newblock \emph{Technology and disability}, 18(4):207--216.

\bibitem[{He et~al.(2016)He, Xia, Qin, Wang, Yu, Liu, and Ma}]{dual1}
Di~He, Yingce Xia, Tao Qin, Liwei Wang, Nenghai Yu, Tie-Yan Liu, and Wei-Ying
  Ma. 2016.
\newblock Dual learning for machine translation.
\newblock \emph{Advances in neural information processing systems}, 29.

\bibitem[{He et~al.(2023)He, Liang, Jiao, Zhang, Yang, Wang, Tu, Shi, and
  Wang}]{he2023exploring}
Zhiwei He, Tian Liang, Wenxiang Jiao, Zhuosheng Zhang, Yujiu Yang, Rui Wang,
  Zhaopeng Tu, Shuming Shi, and Xing Wang. 2023.
\newblock \href {http://arxiv.org/abs/2305.04118} {Exploring human-like
  translation strategy with large language models}.

\bibitem[{He et~al.(2022{\natexlab{a}})He, Wang, Tu, Shi, and
  Wang}]{he-etal-2022-tencent}
Zhiwei He, Xing Wang, Zhaopeng Tu, Shuming Shi, and Rui Wang.
  2022{\natexlab{a}}.
\newblock \href {https://aclanthology.org/2022.wmt-1.18} {Tencent {AI} lab -
  shanghai jiao tong university low-resource translation system for the {WMT}22
  translation task}.
\newblock In \emph{Proceedings of the Seventh Conference on Machine Translation
  (WMT)}, pages 260--267, Abu Dhabi, United Arab Emirates (Hybrid). Association
  for Computational Linguistics.

\bibitem[{He et~al.(2022{\natexlab{b}})He, Wang, Wang, Shi, and
  Tu}]{he-etal-2022-bridging}
Zhiwei He, Xing Wang, Rui Wang, Shuming Shi, and Zhaopeng Tu.
  2022{\natexlab{b}}.
\newblock \href {https://doi.org/10.18653/v1/2022.acl-long.456} {Bridging the
  data gap between training and inference for unsupervised neural machine
  translation}.
\newblock In \emph{Proceedings of the 60th Annual Meeting of the Association
  for Computational Linguistics (Volume 1: Long Papers)}, pages 6611--6623,
  Dublin, Ireland. Association for Computational Linguistics.

\bibitem[{Hsu et~al.(2021)Hsu, Bolte, Tsai, Lakhotia, Salakhutdinov, and
  Mohamed}]{aligner8}
Wei-Ning Hsu, Benjamin Bolte, Yao-Hung~Hubert Tsai, Kushal Lakhotia, Ruslan
  Salakhutdinov, and Abdelrahman Mohamed. 2021.
\newblock Hubert: Self-supervised speech representation learning by masked
  prediction of hidden units.
\newblock \emph{IEEE/ACM Transactions on Audio, Speech, and Language
  Processing}, 29:3451--3460.

\bibitem[{Inan et~al.(2022)Inan, Zhong, Hassan, Quandt, and
  Alikhani}]{modelIntensity}
Mert Inan, Yang Zhong, Sabit Hassan, Lorna Quandt, and Malihe Alikhani. 2022.
\newblock Modeling intensification for sign language generation: A
  computational approach.
\newblock In \emph{Findings of the Association for Computational Linguistics:
  ACL 2022}, pages 2897--2911.

\bibitem[{Kan et~al.(2022)Kan, Hu, Hagenbuchner, Tsoi, Bennamoun, and
  Wang}]{kan2022sign}
Jichao Kan, Kun Hu, Markus Hagenbuchner, Ah~Chung Tsoi, Mohammed Bennamoun, and
  Zhiyong Wang. 2022.
\newblock Sign language translation with hierarchical spatio-temporal graph
  neural network.
\newblock In \emph{Proceedings of the IEEE/CVF Winter Conference on
  Applications of Computer Vision}, pages 3367--3376.

\bibitem[{Karpouzis et~al.(2007)Karpouzis, Caridakis, Fotinea, and
  Efthimiou}]{aligner2}
Kostas Karpouzis, George Caridakis, S-E Fotinea, and Eleni Efthimiou. 2007.
\newblock Educational resources and implementation of a greek sign language
  synthesis architecture.
\newblock \emph{Computers \& Education}, 49(1):54--74.

\bibitem[{Kenton and Toutanova(2019)}]{kenton2019bert}
Jacob Devlin Ming-Wei~Chang Kenton and Lee~Kristina Toutanova. 2019.
\newblock Bert: Pre-training of deep bidirectional transformers for language
  understanding.
\newblock In \emph{Proceedings of NAACL-HLT}, pages 4171--4186.

\bibitem[{Lample et~al.()Lample, Conneau, Denoyer, and
  Ranzato}]{lampleunsupervised}
Guillaume Lample, Alexis Conneau, Ludovic Denoyer, and Marc'Aurelio Ranzato.
\newblock Unsupervised machine translation using monolingual corpora only.
\newblock In \emph{International Conference on Learning Representations}.

\bibitem[{Lin et~al.(2023)Lin, Wang, Zhu, Sun, Zhang, and Yang}]{lin2023gloss}
Kezhou Lin, Xiaohan Wang, Linchao Zhu, Ke~Sun, Bang Zhang, and Yi~Yang. 2023.
\newblock Gloss-free end-to-end sign language translation.
\newblock \emph{arXiv preprint arXiv:2305.12876}.

\bibitem[{Liu et~al.(2017)Liu, Breuel, and Kautz}]{liu2017unsupervised}
Ming-Yu Liu, Thomas Breuel, and Jan Kautz. 2017.
\newblock Unsupervised image-to-image translation networks.
\newblock \emph{Advances in neural information processing systems}, 30.

\bibitem[{Luo et~al.(2017)Luo, Wang, Lin, and Wang}]{dual5}
Ping Luo, Guangrun Wang, Liang Lin, and Xiaogang Wang. 2017.
\newblock Deep dual learning for semantic image segmentation.
\newblock In \emph{Proceedings of the IEEE international conference on computer
  vision}, pages 2718--2726.

\bibitem[{McDonald et~al.(2016)McDonald, Wolfe, Schnepp, Hochgesang, Jamrozik,
  Stumbo, Berke, Bialek, and Thomas}]{aligner3}
John McDonald, Rosalee Wolfe, Jerry Schnepp, Julie Hochgesang, Diana~Gorman
  Jamrozik, Marie Stumbo, Larwan Berke, Melissa Bialek, and Farah Thomas. 2016.
\newblock An automated technique for real-time production of lifelike
  animations of american sign language.
\newblock \emph{Universal Access in the Information Society}, 15:551--566.

\bibitem[{Momeni et~al.(2022)Momeni, Bull, Prajwal, Albanie, Varol, and
  Zisserman}]{momeni2022automatic}
Liliane Momeni, Hannah Bull, KR~Prajwal, Samuel Albanie, G{\"u}l Varol, and
  Andrew Zisserman. 2022.
\newblock Automatic dense annotation of large-vocabulary sign language videos.
\newblock In \emph{European Conference on Computer Vision}, pages 671--690.
  Springer.

\bibitem[{M{\"u}ller et~al.(2022{\natexlab{a}})M{\"u}ller, Ebling, Avramidis,
  Battisti, Berger, Bowden, Braffort, Camg{\"o}z, Espa{\~n}a-Bonet,
  Grundkiewicz et~al.}]{muller2022findings}
Mathias M{\"u}ller, Sarah Ebling, Eleftherios Avramidis, Alessia Battisti,
  Mich{\`e}le Berger, Richard Bowden, Annelies Braffort, Necati~Cihan
  Camg{\"o}z, Cristina Espa{\~n}a-Bonet, Roman Grundkiewicz, et~al.
  2022{\natexlab{a}}.
\newblock Findings of the first wmt shared task on sign language translation
  (wmt-slt22).
\newblock In \emph{Proceedings of the Seventh Conference on Machine Translation
  (WMT)}, pages 744--772.

\bibitem[{M{\"u}ller et~al.(2022{\natexlab{b}})M{\"u}ller, Ebling, Avramidis,
  Battisti, Berger, Bowden, Braffort, Camg{\"o}z, Espa{\~n}a-Bonet,
  Grundkiewicz et~al.}]{WMT2022findings}
Mathias M{\"u}ller, Sarah Ebling, Eleftherios Avramidis, Alessia Battisti,
  Mich{\`e}le Berger, Richard Bowden, Annelies Braffort, Necati~Cihan
  Camg{\"o}z, Cristina Espa{\~n}a-Bonet, Roman Grundkiewicz, et~al.
  2022{\natexlab{b}}.
\newblock Findings of the first wmt shared task on sign language translation
  (wmt-slt22).
\newblock In \emph{Proceedings of the Seventh Conference on Machine Translation
  (WMT)}, pages 744--772.

\bibitem[{Papineni et~al.(2002)Papineni, Roukos, Ward, and Zhu}]{bleu}
Kishore Papineni, Salim Roukos, Todd Ward, and Wei-Jing Zhu. 2002.
\newblock Bleu: a method for automatic evaluation of machine translation.
\newblock In \emph{Proceedings of the 40th annual meeting of the Association
  for Computational Linguistics}, pages 311--318.

\bibitem[{Radford et~al.(2021)Radford, Kim, Hallacy, Ramesh, Goh, Agarwal,
  Sastry, Askell, Mishkin, Clark et~al.}]{radford2021learning}
Alec Radford, Jong~Wook Kim, Chris Hallacy, Aditya Ramesh, Gabriel Goh,
  Sandhini Agarwal, Girish Sastry, Amanda Askell, Pamela Mishkin, Jack Clark,
  et~al. 2021.
\newblock Learning transferable visual models from natural language
  supervision.
\newblock In \emph{International conference on machine learning}, pages
  8748--8763. PMLR.

\bibitem[{Saunders et~al.(2020{\natexlab{a}})Saunders, Camgoz, and
  Bowden}]{aligner5}
Ben Saunders, Necati~Cihan Camgoz, and Richard Bowden. 2020{\natexlab{a}}.
\newblock Adversarial training for multi-channel sign language production.
\newblock \emph{arXiv preprint arXiv:2008.12405}.

\bibitem[{Saunders et~al.(2020{\natexlab{b}})Saunders, Camgoz, and
  Bowden}]{aligner4}
Ben Saunders, Necati~Cihan Camgoz, and Richard Bowden. 2020{\natexlab{b}}.
\newblock Progressive transformers for end-to-end sign language production.
\newblock In \emph{Computer Vision--ECCV 2020: 16th European Conference,
  Glasgow, UK, August 23--28, 2020, Proceedings, Part XI 16}, pages 687--705.
  Springer.

\bibitem[{Sennrich et~al.(2016)Sennrich, Haddow, and
  Birch}]{sennrich2016improving}
Rico Sennrich, Barry Haddow, and Alexandra Birch. 2016.
\newblock Improving neural machine translation models with monolingual data.
\newblock In \emph{ACL}.

\bibitem[{Shi et~al.(2022)Shi, Brentari, Shakhnarovich, and Livescu}]{openasl}
Bowen Shi, Diane Brentari, Gregory Shakhnarovich, and Karen Livescu. 2022.
\newblock Open-domain sign language translation learned from online video.
\newblock In \emph{Proceedings of the 2022 Conference on Empirical Methods in
  Natural Language Processing}, pages 6365--6379.

\bibitem[{Sincan et~al.(2023)Sincan, Camgoz, and Bowden}]{sincan2023context}
Ozge~Mercanoglu Sincan, Necati~Cihan Camgoz, and Richard Bowden. 2023.
\newblock Is context all you need? scaling neural sign language translation to
  large domains of discourse.
\newblock In \emph{Proceedings of the IEEE/CVF International Conference on
  Computer Vision}, pages 1955--1965.

\bibitem[{Song et~al.(2019)Song, Tan, Qin, Lu, and Liu}]{song2019mass}
Kaitao Song, Xu~Tan, Tao Qin, Jianfeng Lu, and Tie-Yan Liu. 2019.
\newblock Mass: Masked sequence to sequence pre-training for language
  generation.
\newblock In \emph{International Conference on Machine Learning}, pages
  5926--5936. PMLR.

\bibitem[{Sutton-Spence and Woll(1999)}]{BSLlinguistics}
Rachel Sutton-Spence and Bencie Woll. 1999.
\newblock \emph{The linguistics of British Sign Language: an introduction}.
\newblock Cambridge University Press.

\bibitem[{Taylor et~al.(2012)Taylor, Mahler, Theobald, and Matthews}]{aligner6}
Sarah~L Taylor, Moshe Mahler, Barry-John Theobald, and Iain Matthews. 2012.
\newblock Dynamic units of visual speech.
\newblock In \emph{Proceedings of the 11th ACM SIGGRAPH/Eurographics conference
  on Computer Animation}, pages 275--284.

\bibitem[{Unterthiner et~al.(2019)Unterthiner, van Steenkiste, Kurach,
  Marinier, Michalski, and Gelly}]{fvd}
Thomas Unterthiner, Sjoerd van Steenkiste, Karol Kurach, Rapha{\"e}l Marinier,
  Marcin Michalski, and Sylvain Gelly. 2019.
\newblock Fvd: A new metric for video generation.

\bibitem[{Wang et~al.(2023)Wang, Chen, Wu, Zhang, Zhou, Liu, Chen, Liu, Wang,
  Li et~al.}]{aligner9}
Chengyi Wang, Sanyuan Chen, Yu~Wu, Ziqiang Zhang, Long Zhou, Shujie Liu, Zhuo
  Chen, Yanqing Liu, Huaming Wang, Jinyu Li, et~al. 2023.
\newblock Neural codec language models are zero-shot text to speech
  synthesizers.
\newblock \emph{arXiv preprint arXiv:2301.02111}.

\bibitem[{Xia et~al.(2017{\natexlab{a}})Xia, Bian, Qin, Yu, and Liu}]{dual3}
Yingce Xia, Jiang Bian, Tao Qin, Nenghai Yu, and Tie-Yan Liu.
  2017{\natexlab{a}}.
\newblock Dual inference for machine learning.
\newblock In \emph{IJCAI}, pages 3112--3118.

\bibitem[{Xia et~al.(2017{\natexlab{b}})Xia, Qin, Chen, Bian, Yu, and
  Liu}]{dual2}
Yingce Xia, Tao Qin, Wei Chen, Jiang Bian, Nenghai Yu, and Tie-Yan Liu.
  2017{\natexlab{b}}.
\newblock Dual supervised learning.
\newblock In \emph{International conference on machine learning}, pages
  3789--3798. PMLR.

\bibitem[{Xia et~al.(2018)Xia, Tan, Tian, Qin, Yu, and Liu}]{dual6}
Yingce Xia, Xu~Tan, Fei Tian, Tao Qin, Nenghai Yu, and Tie-Yan Liu. 2018.
\newblock Model-level dual learning.
\newblock In \emph{International Conference on Machine Learning}, pages
  5383--5392. PMLR.

\bibitem[{Yan et~al.(2021)Yan, Zhang, Abbeel, and Srinivas}]{videogpt}
Wilson Yan, Yunzhi Zhang, Pieter Abbeel, and Aravind Srinivas. 2021.
\newblock Videogpt: Video generation using vq-vae and transformers.
\newblock \emph{arXiv preprint arXiv:2104.10157}.

\bibitem[{Ye et~al.(2022)Ye, Jiao, Wang, and Tu}]{ye2022scaling}
Jinhui Ye, Wenxiang Jiao, Xing Wang, and Zhaopeng Tu. 2022.
\newblock Scaling back-translation with domain text generation for sign
  language gloss translation.
\newblock \emph{arXiv preprint arXiv:2210.07054}.

\bibitem[{Ye et~al.(2023)Ye, Jiao, Wang, Tu, and Xiong}]{ye2023cross}
Jinhui Ye, Wenxiang Jiao, Xing Wang, Zhaopeng Tu, and Hui Xiong. 2023.
\newblock Cross-modality data augmentation for end-to-end sign language
  translation.
\newblock \emph{arXiv}.

\bibitem[{Yi et~al.(2017)Yi, Zhang, Tan, and Gong}]{dual4}
Zili Yi, Hao Zhang, Ping Tan, and Minglun Gong. 2017.
\newblock Dualgan: Unsupervised dual learning for image-to-image translation.
\newblock In \emph{Proceedings of the IEEE international conference on computer
  vision}, pages 2849--2857.

\bibitem[{Yin et~al.(2021)Yin, Moryossef, Hochgesang, Goldberg, and
  Alikhani}]{yin2021including}
Kayo Yin, Amit Moryossef, Julie Hochgesang, Yoav Goldberg, and Malihe Alikhani.
  2021.
\newblock Including signed languages in natural language processing.
\newblock \emph{arXiv preprint arXiv:2105.05222}.

\bibitem[{Yin and Read(2020)}]{yin2020better}
Kayo Yin and Jesse Read. 2020.
\newblock Better sign language translation with stmc-transformer.
\newblock In \emph{Proceedings of the 28th International Conference on
  Computational Linguistics}, pages 5975--5989.

\bibitem[{Zhang et~al.()Zhang, M{\"u}ller, and Sennrich}]{sltunet}
Biao Zhang, Mathias M{\"u}ller, and Rico Sennrich.
\newblock Sltunet: A simple unified model for sign language translation.
\newblock In \emph{International Conference on Learning Representations}.

\bibitem[{Zhang et~al.(2023)Zhang, Li, and Bing}]{videoLLama}
Hang Zhang, Xin Li, and Lidong Bing. 2023.
\newblock Video-llama: An instruction-tuned audio-visual language model for
  video understanding.
\newblock \emph{arXiv preprint arXiv:2306.02858}.

\bibitem[{Zhang et~al.(2022)Zhang, Yuan, Liao, and Zhang}]{aligner7}
Sibo Zhang, Jiahong Yuan, Miao Liao, and Liangjun Zhang. 2022.
\newblock Text2video: Text-driven talking-head video synthesis with
  personalized phoneme-pose dictionary.
\newblock In \emph{ICASSP 2022-2022 IEEE International Conference on Acoustics,
  Speech and Signal Processing (ICASSP)}, pages 2659--2663. IEEE.

\bibitem[{Zhou et~al.(2021{\natexlab{a}})Zhou, Zhou, Qi, Pu, and Li}]{signBT}
Hao Zhou, Wengang Zhou, Weizhen Qi, Junfu Pu, and Houqiang Li.
  2021{\natexlab{a}}.
\newblock Improving sign language translation with monolingual data by sign
  back-translation.
\newblock In \emph{Proceedings of the IEEE/CVF Conference on Computer Vision
  and Pattern Recognition}, pages 1316--1325.

\bibitem[{Zhou et~al.(2021{\natexlab{b}})Zhou, Zhou, Zhou, and
  Li}]{zhou2021spatial}
Hao Zhou, Wengang Zhou, Yun Zhou, and Houqiang Li. 2021{\natexlab{b}}.
\newblock Spatial-temporal multi-cue network for sign language recognition and
  translation.
\newblock \emph{IEEE Transactions on Multimedia}, 24:768--779.

\end{thebibliography}
\bibliographystyle{acl_natbib}

\appendix
\section{APPENDIX}
    \label{tab:order}

\subsection{QULITATIVE RESULTS AND FAILURE ANALYSIS}
Overall the results in Table ~\ref{tab:slt}  are seemingly  poor in BOBSL dataset. We dig deep into 'why' the results are poor and to work towards building an understanding for "how" they can be improved significantly. 

\paragraph{Regarding the "Why" Aspect} We conduct a thorough  analysis of the results, identifying the areas in which our approach performs well and those that require further improvement. 

Initially, we conduct thorough case study including good cases, bad cases and comparision case between USLNet (unsupervised setting ) and ~\citet{bbc} which is one supervised model.  From digging into our results in Table ~\ref{tab:caseStudy} , we find that we can do relatively better in Main ingredients (eg: bus, I, anything), but always fail in other detail, such as proper noun (eg: Ma Effanga), and complex sentence (which is that).


\begin{table*}[tbp]
\centering
\resizebox{\linewidth}{!}{
\begin{tabular}{ll}
\toprule
Golden Text: & It’s quite a journey \textbf{especially} if \textbf{I get the bus}. \\
USLNet: & It’s \textbf{especially} long if I \textbf{get the bus}.  \\ 
\hline
Golden Text: &  It’s hell of a \textbf{difference} yeah.\\
USLNet: & It’s \textbf{different completely}. \\ 

\hline
Golden Text: & Oh, \textbf{Ma Effanga} is going to be green.\\
USLNet: & It’s not going to be green.\\ 
\hline
Golden Text:  & They started challenging the sultan  in a very important aspect, \\ 
&\textbf{which is that he is not Muslim enough}.\\
USLNet: & This is a very important aspect.\\ 

\hline
Golden Text: & It’s quite a journey  \textbf{especially}  if \textbf{I get the bus}. \\
USLNet: & It’s \textbf{especially}  long if \textbf{I get the bus}. \\ 
\citet{bbc}: & How long have you been in the \textbf{bus} now. \\ 

\hline
Golden Text: & It’s hell of  a \textbf{difference} yeah. \\
USLNet:& It’s \textbf{different completely}.\\ 
\citet{bbc}: & It was like trying  to be \textbf{different} to the world. \\

\bottomrule
\end{tabular}}
\caption{Case study for USLNet on BOBSL. Examples are from the test set. There are six examples. The first and second examples show the relatively good cases decoded by USLNet. The third and fourth cases show the shortcomings of USLNet. The last two cases show our unsupervised model obtain competive examples comparing with supervised model.}
\label{tab:caseStudy}
\end{table*}

What's more, the comparision case between USLNet (unsupervised setting ) and \citet{bbc} is as follows. From the Table ~\ref{tab:caseStudy} , we observe that our outcomes are competitive with those of supervised methods. Furthermore, in certain instances, we can achieve more accurate output (for example, particularly in specific cases).



\paragraph{Regarding the "How" Aspect}  We propose a two-fold approach. Firstly, we suggest allowing unsupervised learning to serve as a representation learning stage. From the  Table ~\ref{tab:slt}, we can use unuspervised training way can provide one good representation and is significant for improve supervised translation method, resulting in a substantial increase in the BLEU-4 score from 1.0 to 1.4. Secondly, we recommend enhancing USLNet by focusing on improvements in both the pretraining and aligner components.

USLNet can be divided into two primary components: the pretraining module (comprising the text pre-training module and the video pre-training module) and the mapper part (slide window aligner). Consequently, the paths to success can be categorized into two aspects. The first aspect involves pre-training, where we can adapt our method using multi-modal models, such as videoLLama ~\citep{videoLLama}. The second aspect focuses on designing an effective mapper~\citep{aligner4,aligner5}.

\subsection{DIFFERENT ALIGNMENT NETWORKS}
 The effects of different alignment networks for sign language generation are in Table ~\ref{tab:slgDifferent}. We can observe that our method achieves the best performance, demostrating the effectiveness of USLNet.
 \begin{table}[htbp]
    \centering
    \setlength{\tabcolsep}{3pt}
    \begin{tabular}{l | c  | cc}
        \toprule
         \multirow{2}{*}{\bf Method}  &  \multicolumn{1}{c|}{\bf Dev } &  \multicolumn{2}{c}{\bf Test} \\
        & B@1 $\uparrow$  & B@1 $\uparrow$ & B@4$\uparrow$  \\
        \midrule
        Pooling                              & 7.00   &  6.60   & 0.00  \\
        Sliding Window Aligner    & 11.70  &  11.70  & 0.00 \\ 
        (hard connection) &&&\\
         Sliding Window Aligner  & 20.90  &  22.70  & 0.20 \\
         (soft connection) &&&\\
        \bottomrule
    \end{tabular}
     \caption{Sign language generation results in terms of back-translation BLEU of USLNet with different cross-modality mappers on BOBSL. B@1 and denotes BLEU-1 and BLEU-4, respectively.}
    \label{tab:slgDifferent}
\end{table} 

\subsection{ADDITIONAL RELATED WORK}
\paragraph{Text-to-Video Aligner}  Text-to-video aligners in sign language domain can be broadly classified into two main categories. The first category involves the use of animated avatars to generate sign language, relying on a predefined text-sign dictionary that converts text phrases into sign pose sequences ~\citep{aligner1,aligner2,aligner3}. The second category encompasses deep learning approaches applied to text-video mapping. ~\citet{aligner4,aligner5} adapt the transformer architecture to the text-video domain and employ a linear embedding layer to map the visual embedding into the corresponding space. Unlike these methods, which can only decode pose images, our Unsupervised Sequence Learning Network (USLNet) is capable of generating videos. We address the length and dimension mismatch issues by utilizing a simple sliding window aligner.

Text-to-video aligners in other domains have also been proposed. ~\citet{aligner6} introduced a method for automatic redubbing of videos by leveraging the many-to-many mapping of phoneme sequences to lip movements, modeled as dynamic visemes. The Text2Video approach ~\citet{aligner7} employs a phoneme-to-pose dictionary to generate key poses and high-quality videos from phoneme-poses. This phoneme-pose dictionary can be considered as a token-token mapper. Analogously, USLNet quantizes discrete videos and extracts video tokens, a standard technique in the audio domain ~\citep{aligner8,aligner9,aligner10}. Consequently, the sliding window aligner also serves as a token-token aligner. However, unlike the Text2Video method, which performs a lookup action to obtain target tokens, our approach decodes the target token using all source tokens.

\paragraph{Dual Learning}   ~\citet{dual1} propose dual learning to reduce the requirement on labeled data aiming to train English-to-French and French-to-English translators. It regards that French-to-English translation is the dual task to English-to-French translation. Thus, it designs to set up a dual-learning game which two agents , each of whom only understands one language and can evaluate how likely the translated are natural sentences in targeted language and to what to extent the reconstructed are consistent with the original. Moreover, researchers exploit the duality between two tasks in training~\citep{dual2} and inference ~\citep{dual3} stage , so as to achieve better performance. Dual learning algorithms have been proposed for different tasks, such as translation~\citep{dual1}, sentence analysis~\citep{dual6}, image-image translation~\citep{dual4}, image segmentation~\citep{dual5}. USLNet extend dual learning to sign language realm and design dual cross-modality back-translation to learn sign language translation and generation tasks in one unified way.

\subsection{ADDITIONAL ANALYSIS}
\paragraph{Mass text pretraining method outperform than Mlm method} In this study, we conduct a comparative analysis of various text pretraining methods to assess their impact on sign language translation task shown in Table ~\ref{tab:caseStudy}. Specifically, we focus on comparing the performance of the masked language modeling (MLM) ~\citep{kenton2019bert}  method and the recently proposed masked sequence-to-sequence (Mass) ~\citep{song2019mass}. Our findings reveal that the MASS method outperforms the MLM method (+1.00 BLEU-1 score) in terms of enhancing the model's ability to capture semantic relationships and improve the overall quality of the learned representations.
\paragraph{Multi-task modeling benefits SLT} 
Multi-task modeling in sign language translation (SLT) presents significant advantages. The incorporation of multiple tasks, particularly the inclusion of cross-modality back-translation, within the modeling framework allows SLT systems to leverage shared representations and tap into a diverse range of informational sources. Our empirical analysis, as depicted in Table ~\ref{tab:caseStudy}, substantiates the meaningful impact of key components on SLT performance. Specifically, our findings demonstrate a substantial decrease in SLT results when text reconstruction is omitted (-3.2 BLEU-1 score), video reconstruction is absent (-1.3 BLEU-1 score), or cross-modality back-translation training is neglected (-9 BLEU-1 score). These observations underscore the crucial role of these components in achieving optimal performance in SLT.

\begin{table*}[h]
\centering
\begin{tabular}{cccc}
\toprule
ID & System & SLT B@1$^{\text{↑}}$ \\
\midrule
1 & Baseline & 3.20  \\
1.1 & 1+more text data & 9.60 \\
\hline
Adjust data distribution & &  \\
2 & 1.1+ 1M parallel video and text  & \\
& for unsupervised training & 15.20  \\
\hline
Explore Different text pretraining method & &  \\
3.1 & 2+ mlm text pretrain method  & 15.20  \\
3.2 & 2+ mass text pretrain & 16.20 \\
\bottomrule
\end{tabular}
\label{tab:moreAblation}
\caption{Additional Ablation study of UnSLNet on sign language translation(SLT) on the BOBSL dev set.}
\end{table*}

\subsection{DISCUSSION ABOUT \citet{bbc}.}
In terms of model architecture, both Albanie 2021 and USLNet employ a standard transformer encoder-decoder structure. In the Albanie method, the encoder and decoder comprise two attention layers, each with two heads. Conversely, USLNet adopts a large model architecture, setting the encoder and decoder layers to six. Regarding methodology, Albanie 2021 utilizes a supervised approach for learning sign language translation. In contrast, USLNet employs an unsupervised method, leveraging an abundant text corpus to learn text generation capabilities and employing video-text-video back-translation to acquire cross-modality skills. Concerning model output, Albanie 2021 has released several qualitative examples. We have compared these with the results from USLNet, which demonstrate that USLNet achieves competitive outcomes in comparison to the supervised method.
\newpage

\subsection{QUALITATIVE VISUAL RESULTS}

\begin{figure}[h]
	\centering
	\includegraphics[width=\columnwidth]{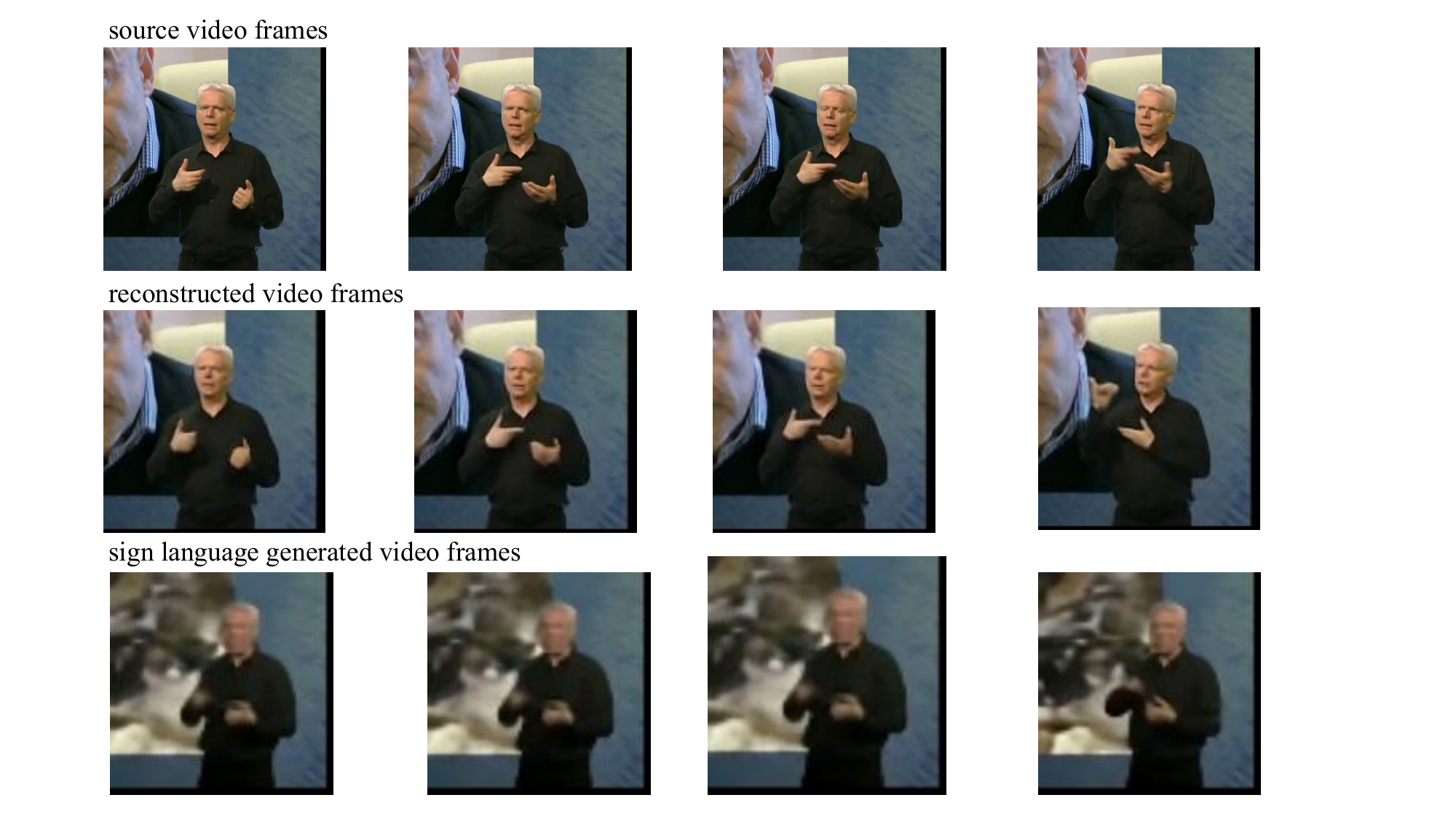}
	\caption{Case study of UnSLNet on BOBSL for sign language generation task. Examples are from test set.}
	\label{fig:{pic/model}}
\end{figure}

\begin{figure}[h]
	\centering
	\includegraphics[width=\columnwidth]{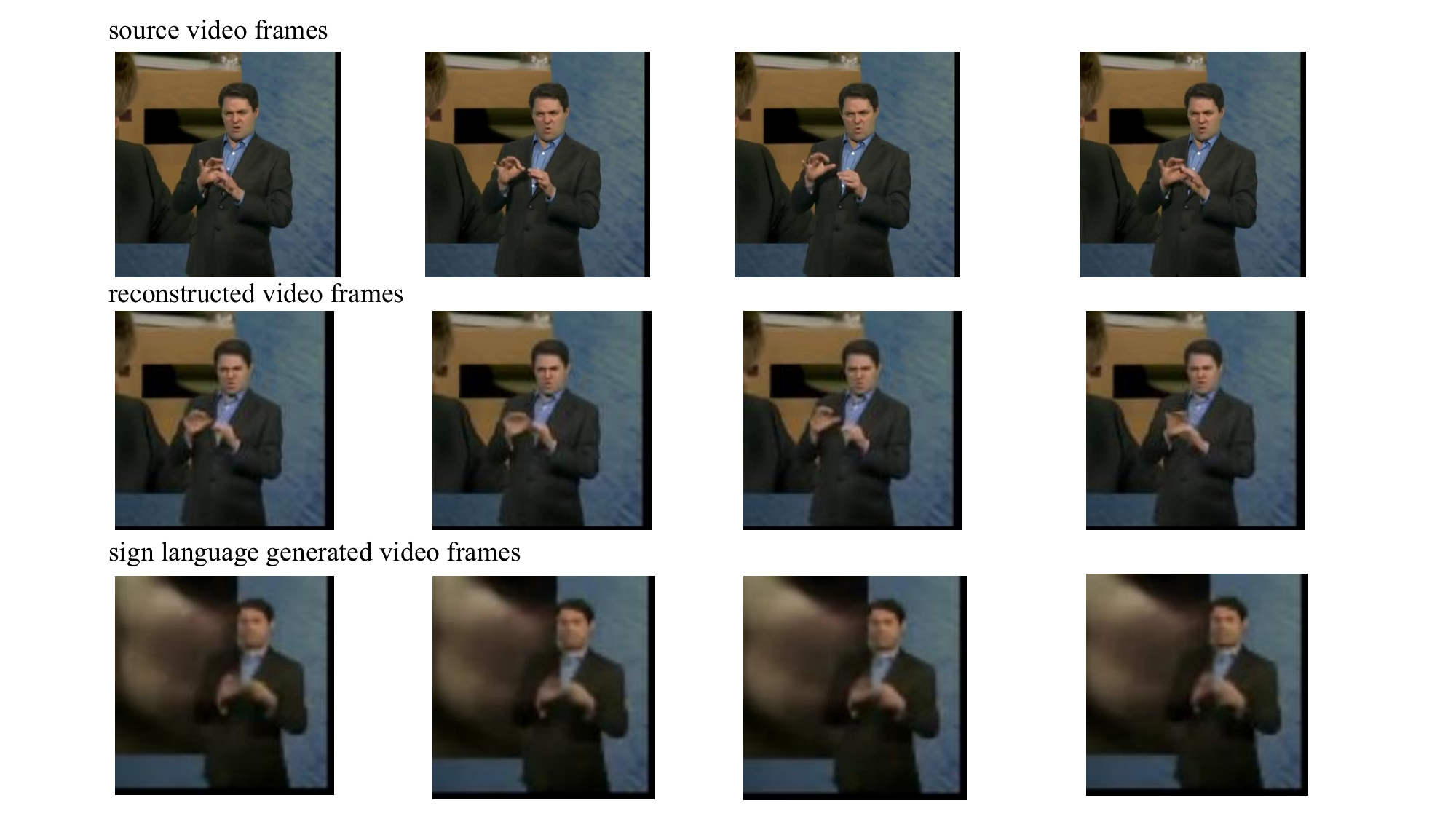}
	\caption{Case study of UnSLNet on BOBSL for sign language generation task. Examples are from test set.}
	\label{fig:{pic/model}}
\end{figure}





\end{document}